\documentclass{article}

\usepackage{arxiv}

\usepackage[utf8]{inputenc} 
\usepackage[T1]{fontenc}    
\usepackage{hyperref}       
\usepackage{url}            
\usepackage{booktabs}       
\usepackage{amsfonts}       
\usepackage{nicefrac}       
\usepackage{threeparttable}
\usepackage{microtype}      
\usepackage{lipsum}
\usepackage{graphicx}
\usepackage{hyperref}
\usepackage{epstopdf, epsfig}
\usepackage{bm}
\usepackage{amsthm}
\usepackage{amsmath}
\usepackage[scr=esstix]{mathalpha}
\usepackage{amssymb}
\usepackage[utf8]{inputenc}
\usepackage{multirow,booktabs}
\usepackage{tabularx}
\usepackage[T1]{fontenc}
\usepackage{mathptmx}
\usepackage{multirow}
\usepackage{algorithm}
\usepackage{algpseudocode} 
\usepackage[numbers]{natbib}
\usepackage{float}
\usepackage{eucal}
\usepackage{color}
\usepackage{subcaption}
\usepackage{natbib}
\usepackage{caption}
\title{\normalfont Scalable $h$-adaptive probabilistic solver for time-independent and time-dependent systems}

\author{
  Akshay Thakur \thanks{These authors contributed equally to this work.}\\
  Department of Aerospace and Mechanical Engineering\\
  University of Notre Dame\\
  Notre Dame, IN-46556, USA.\\
  \texttt{athakur3@nd.edu} \\
  \And
  Sawan Kumar\footnotemark[1] \\
  Department of Applied Mechanics\\
  Indian Institute of Technology Delhi\\
  Hauz Khas - 110016, New Delhi, India \\
  \texttt{sawan.kumar@am.iitd.ac.in} \\
  \And
   Matthew Zahr \\
  Department of Aerospace and Mechanical Engineering\\
  University of Notre Dame\\
  Notre Dame, IN-46556, USA.\\
  \texttt{mzahr@nd.edu} \\
   \And
 Souvik Chakraborty \\
  Department of Applied Mechanics\\
  School of Artificial Intelligence\\
  Indian Institute of Technology Delhi\\
  Hauz Khas - 110016, New Delhi, India \\
  \texttt{souvik@am.iitd.ac.in} \\
}

\begin{document}
\maketitle

\begin{abstract}
Solving partial differential equations (PDEs) within the framework of probabilistic numerics offers a principled approach to quantifying epistemic uncertainty arising from discretization. By leveraging Gaussian process regression and imposing the governing PDE as a constraint at a finite set of collocation points, probabilistic numerics delivers mesh-free solutions at arbitrary locations. However, the high computational cost, which scales cubically with the number of collocation points, remains a critical bottleneck, particularly for large-scale or high-dimensional problems. We propose a scalable enhancement to this paradigm through two key innovations. First, we employ a stochastic dual descent algorithm that reduces the per-iteration complexity from cubic to linear in the number of collocation points, enabling tractable inference. Second, we exploit a clustering-based active learning strategy that adaptively selects collocation points to maximize information gain while minimizing computational expense. Together, these contributions result in an $h$-adaptive probabilistic solver that can scale to a large number of collocation points. We demonstrate the efficacy of the proposed solver on benchmark PDEs, including two- and three-dimensional steady-state elliptic problems, as well as a time-dependent parabolic PDE formulated in a space-time setting.

\end{abstract}

\section{Introduction}
Partial differential equations (PDEs) are fundamental tools for modeling physical \cite{sommerfeld1949partial}, biological \cite{jones2009differential}, and engineering systems \cite{ames2016nonlinear}, governing phenomena such as heat conduction \cite{bergman2011introduction}, fluid flow \cite{kundu2024fluid}, elasticity \cite{barber2023elasticity}, and diffusion \cite{bergman2011introduction}. Over the decades, the advancement and use of classical numerical methods such as finite difference \cite{leveque1998finite}, finite element \cite{reddy1993introduction}, and finite volume methods \cite{leveque2002finite} have been instrumental in solving PDEs, which in turn has enabled advances in a wide range of fields. However, these traditional methods generally yield deterministic, point-valued solutions and do not inherently capture the uncertainties that arise from discretization errors. In high-stakes applications such as weather forecasting \cite{bauer2015quiet}, structural health diagnostics \cite{farrar2007introduction}, and patient-specific biomedical simulations \cite{taylor2009patient}, neglecting these sources of uncertainty can lead to misleading predictions and potentially flawed decision-making. As such, developing probabilistic solvers is increasingly being recognized as critical for building reliable, interpretable, and robust computational models in complex real-world scenarios \cite{kumar2025towards,kramer2024stable,hennig2015probabilistic,hennig2015probabilisticnum,bilionis2016probabilistic,cockayne2017probabilistic,chen2021solving}.

Probabilistic numerics \cite{hennig2015probabilisticnum} represents a transformative paradigm that reinterprets traditional numerical computation through the lens of statistical inference. Rather than producing only deterministic solutions, probabilistic numerics-based approaches treat the solution to a computational problem, such as a PDE, as a latent function to be inferred under uncertainty. In the context of PDEs, this means constructing a probabilistic model that integrates the prior knowledge about the structure of the solution with observed information on how well it satisfies the governing equations. A common strategy within this framework involves placing a Gaussian process (GP) \cite{bilionis2016probabilistic,chen2021solving,cockayne2017probabilistic,pfortner2022physics} prior over the unknown solution field. The GP captures beliefs about smoothness and spatial correlation, while residuals of the PDE evaluated at selected collocation points are treated as noisy observations. Conditioning the prior on these residuals yields a posterior distribution over the solution space. This posterior not only provides an estimate of the solution but also quantifies epistemic uncertainty arising from limited collocation points. One of the key advantages of this approach is its mesh-free nature: it avoids the need for structured grids or mesh generation, which are often computationally expensive and difficult to construct in complex geometries \cite{liu2009meshfree}. Overall, by unifying solution estimation with uncertainty quantification, probabilistic numerics opens new avenues for robust and interpretable modeling of complex systems.


Despite the significant promise of GP-based probabilistic solvers for PDEs, their practical deployment has been limited by two major challenges:
\begin{itemize}
    \item \textbf{Scalability} is a fundamental bottleneck arising from the computational structure of Gaussian process inference. Specifically, GP models require the inversion of dense kernel (covariance) matrices whose size is determined by the number of collocation points $N$ used to enforce the PDE constraints. This operation has a computational cost of $\mathcal O \left(N^3\right)$ and a memory cost of $\mathcal O \left(N^2\right)$, making it prohibitively expensive for large-scale simulations. As a result, GP-based solvers struggle to scale to high-resolution spatial discretizations, high-dimensional domains, or long-time dynamical problems—scenarios where conventional numerical solvers excel.
    \item \textbf{Adaptivity} presents an additional challenge. The placement of collocation points affects the solver’s ability to capture important solution features. Uniform or random sampling often leads to poor resolution in regions with steep gradients, sharp interfaces, or localized phenomena. While classical numerical methods, such as finite elements, incorporate adaptive refinement strategies (e.g., $h$-adaptivity, where the mesh is refined in regions of high error), these strategies are not readily integrated into standard probabilistic numerical frameworks. The lack of principled, data-driven strategies for adaptive point placement in GP-based solvers limits their efficiency and accuracy, particularly in heterogeneous or multiscale problems.
\end{itemize}

To address these limitations, we propose a novel $h$-adaptive, scalable probabilistic solver that enables efficient and accurate solution of PDEs while preserving the core benefits of the probabilistic numeric framework. 
Our work makes a step toward making probabilistic numerics practical and scalable for real-world PDE problems. By integrating ideas from stochastic optimization and adaptive learning into the probabilistic numeric framework, we bridge the gap between principled uncertainty quantification and computational traceability. 
The key features of the proposed solver include
\begin{itemize}
    \item \textbf{Stochastic dual descent for scalable inference}: We propose the utilization of a stochastic dual  (SDD) algorithm for optimizing the representer weights of the Gaussian process posterior. By reformulating the inference problem in the dual space and employing stochastic optimization over mini-batches of collocation points, we reduce the per-iteration cost from cubic to linear in $N$. This makes our approach suitable for large-scale problems where classical GP inference would be computationally infeasible.
    \item \textbf{Clustering-based active learning for $h$-adaptivity}: To mimic $h$-adaptivity and improve data efficiency, we employ a clustering-based active learning strategy for collocation point selection. By leveraging posterior uncertainty and solution features, our method refines the placement of collocation points in regions where the solution is complex or poorly resolved, thereby improving accuracy without unnecessary computational overhead.
    \item We validate our method on a range of benchmark problems, including two- and three-dimensional steady-state elliptic PDEs and a space-time-formulated time-dependent parabolic PDE. These examples demonstrate the scalability, adaptivity, and uncertainty quantification capabilities of our approach across different geometries and problem classes.
\end{itemize}

The remainder of this manuscript is structured as follows: Section \ref{BG}  and Section \ref{Framework} delineate the mathematical underpinnings and provide a comprehensive exposition of the proposed framework. Section \ref{Numercial_eg} presents a detailed demonstration of our approach on three different numerical examples, including two and three-dimensional steady elliptic PDEs and a time-dependent parabolic PDE solved in a space-time setting. Lastly, Section \ref{Conclusion} provides the concluding remarks.

\section{Problem statement}\label{BG}

Consider a linear partial differential equation $\mathcal{P}$ defined on a spatial domain $\Omega \subset \mathbb{R}^d$, given by:
\begin{equation}
\mathcal{L}[u](\bm{x},t) = s(\bm{x},t), \quad \forall \bm{x} \in \Omega,\quad t \in \left(0,T \right]
 \label{eq:pde}
\end{equation}
This equation is subject to linear boundary conditions imposed on the domain boundary $\partial \Omega$ of the domain, expressed as:
\begin{equation}
\mathcal{B}[u](\bm{x}) = b(\bm{x}), \quad \forall \bm{x} \in \partial \Omega,
\label{eq:pdebc}
\end{equation}
where $u \in  \mathbb{R}$ denotes the solution of the partial differential equation, $\mathcal{L}$ represents a linear differential operator acting on the solution within the domain, and $s: \mathbb{R}^{d} \times \mathbb{R} \mapsto \mathbb{R} $ is the source term. Furthermore, $\mathcal{B}$ is a linear differential operator defined on the boundary, and $b:\mathbb{R}^d \mapsto \mathbb{R}$ is the boundary term which specifies the boundary condition. Similarly, the initial condition is represented as:
\begin{equation}
    u\left(\bm x, t=0\right) = u_0,
\end{equation}
where $u_0$ represents the initial condition. 
Note that this formulation reduces to a time-independent system of the following form, in case the dependence on time is not present.
\begin{equation}\label{eq:time_ind_PDE}
    \mathcal L \left[u\right]\left(\bm x \right) = s\left(\bm x \right), \quad \forall \bm x \in \Omega,
\end{equation}
\begin{equation}\label{eq:time_ind_PDE_BC}
    \mathcal B \left[ u \right] \left(\bm x \right) = s \left(\bm x \right), \quad \forall \bm x \in \partial \Omega.
\end{equation}
Conventionally, PDEs of the form in Eqs. \eqref{eq:pde} and \eqref{eq:time_ind_PDE} are solved using numerical techniques such as finite element, finite volume, and finite difference methods.
However, these methods are deterministic and fail to quantify the uncertainty due to discretization. Accordingly, the objective of this work is to develop a computationally efficient probabilistic solver, denoted by $\mathcal{M}$, which can be formally represented as 
\begin{equation}
\mathcal{M} : \mathcal{M}_{X,\tau} \times \Theta \to \mathcal{M}_{\theta}.
\label{eq:solver_corrected_1_latex}
\end{equation}
Here, $\bm{x},t \in \mathcal{M}_{X,\tau}$ represents the input vector, and $\mathcal{M}_{X,\tau}$ is the corresponding spatio-temporal input space. The event space $\Theta$, associated with the probability space $(\Theta, \Sigma, P)$, encapsulates the epistemic uncertainty arising from the construction of the solver. This uncertainty is manifested through solver choices such as the number and distribution of discretization points. $\mathcal{M}_{\theta}$ is a stochastic map induced by the epistemic uncertainty from the input vector to a realization from the epistemic probability measure of the PDE solution.

\section{Methodology}\label{Framework}
This section delineates the proposed framework, providing an exposition of its mathematical underpinnings. Specifically, we elaborate on the essential components, including the GP-based probabilistic solver (GPPS), obtaining the inverse of the kernel Gram matrix with the SDD optimization scheme, the clustering-based active learning strategy employed for the efficient selection of collocation points, and the methodology for sampling from the solver's posterior distribution. However, before that, we remark that since we handle the cubic complexity of solving the PDE using GP in this article, we adopt a space-time formulation instead of employing explicit time-marching schemes for unsteady PDEs. Consequently, we reformulate our PDE $\mathcal{P}$ in Eq. \eqref{eq:pde} by embedding both spatial and temporal dimensions into a unified spatio-temporal domain, denoted by $\Omega_{\mathcal{S}} \subset \mathbb{R}^{d+1}$. We define the corresponding space-time coordinate variable as $\bm{z} = (\bm{x}, t) \in \Omega_{\mathcal{S}}$. It is worth noting that this unifies Eqs. \eqref{eq:pde} and \eqref{eq:time_ind_PDE}, with both being represented as:
\begin{equation}\label{eq:unified_PDE}
    \mathcal L \left[u\right]\left(\bm z \right) = s\left(\bm z \right), \quad \forall \bm z \in \Omega_{\mathcal S}.
\end{equation}
Note that for the steady-state scenario $\Omega_{\mathcal S} \subset \mathbb R^d$, as the time coordinate is absent. Similarly, the boundary conditions can be represented as
\begin{equation}\label{eq:unifiedBCIC}
    \mathcal{B}_c \left[u \right]\left(\bm z \right) = b_c\left(\bm z \right), \quad \forall \bm z \in \partial \Omega_{\mathcal S}
\end{equation}
where $b_c\left(\cdot\right)$ is the space-time boundary term, which includes the initial condition as one of the boundary terms.
The formulation used in the remainder of the article is based on Eqs. \eqref{eq:unified_PDE} and \eqref{eq:unifiedBCIC}.

\subsection{GP-based probabilistic solver}\label{ss:GPPS}
Consider the following conventional GP prior probability measure for solution $u$
\begin{equation}
    u(\bm{z}) \sim \mathcal{GP}(0, k(\bm{z},\bm{z}')),
    \label{eq:prior}
\end{equation}
where $k: \mathbb{R}^{d+1}\times \mathbb{R}^{d+1} \mapsto \mathbb{R}$ is the prior covariance function, where the implicit mapping corresponding to the kernel function from the input space to the feature space $\mathcal{H}$, a Hilbert space, is defined through the following feature map
\begin{equation}
    \phi : \mathbb{R}^{d+1} \mapsto \mathcal{H} \text{ such that } k(\bm{z}, \bm{z}') = \langle \phi(\bm{z}), \phi(\bm{z}') \rangle_{\mathcal{H}}.
\end{equation}

Consider a set of collocation points $\mathcal{Z}_{col} = \mathcal{Z}_{int} \cup \mathcal{Z}_{bnd}$, where $\mathcal{Z}_{int} = \{\bm{z}_i\}_{i=1}^{n_i} \subset \Omega_{\mathcal{S}}$ is a set of $n_i$ collocation point in the interior of the domain and $\mathcal{Z}_{bnd} = \{\bm{z}_j\}_{j=1}^{n_b} $ is the set of $n_b$ collocation points on the domain boundary. 
Leveraging the linearity property of GPs, which dictates that a linear transformation of a Gaussian random variable yields another Gaussian random variable, we constrain the prior probability measure by conditioning on Eqs. \eqref{eq:unified_PDE} and \eqref{eq:unifiedBCIC} evaluated at the set of collocation points $\mathcal{Z}_{col}$. This conditioning process yields a posterior probability measure for the solution $u(\bm{z})$, characterized by the following GP
\begin{equation}
 \left(u|\mathcal{P}\right)(\bm{z}) \sim \mathcal{GP}(\hat{m}_{u|\mathcal{P}}(\bm{z}), \hat{k}_{u|\mathcal{P}}(\bm{z},\bm{z}')).
\end{equation}
Here, the posterior mean function $\hat{m}_{u|\mathcal{P}}(\bm{z})$ and the posterior covariance function $\hat{k}_{u|\mathcal{P}}(\bm{z},\bm{z}')$  are given by
\begin{align}
\hat{m}_{u|\mathcal{P}}(\bm{z}) &= \mathcal{A}(\bm{z})^T \mathbf{K}^{-1}g(\mathcal{Z}_{col}), \\
\hat{k}_{u|\mathcal{P}}(\bm{z},\bm{z}') &= k(\bm{z},\bm{z}') - \mathcal{A}(\bm{z})^T \mathbf{K}^{-1} \mathcal{A}(\bm{z}'), \quad 
\end{align}
where the Gram matrix $\mathbf{K}$, the vector $\mathcal{A}(\bm{z})$, and the vector $g(\mathcal{Z}_{col})$ are defined as
\begin{equation}
 \mathbf{K} =
 \begin{pmatrix}
 \mathcal{L}\mathcal{L}'\bm{k}(\mathcal{Z}_{int}, \mathcal{Z}_{int}) & \mathcal{L}\mathcal{B}'_c\bm{k}(\mathcal{Z}_{int}, \mathcal{Z}_{bnd}) \\
(\mathcal{L}\mathcal{B}'_c\bm{k}(\mathcal{Z}_{int}, \mathcal{Z}_{bnd}))^T & \mathcal{B}_c\mathcal{B}'_c\bm{k}(\mathcal{Z}_{bnd}, \mathcal{Z}_{bnd})
\end{pmatrix}, \quad
\end{equation}
\begin{equation}
\mathcal{A}(\bm{z}) =
\begin{pmatrix}
\mathcal{L}\bm{k}(\bm{z},\mathcal{Z}_{int}) \\
\mathcal{B}_c\bm{k}(\bm{z},\mathcal{Z}_{bnd})
\end{pmatrix}, \quad
\end{equation}
\begin{equation}
g(\mathcal{Z}_{col}) =
\begin{pmatrix}
\bm{s}(\mathcal{Z}_{int}) \\
\bm{b}_{c}(\mathcal{Z}_{bnd})
\end{pmatrix}. \quad
\end{equation}
Here, \( \mathcal{O}\mathcal{O}'\bm{k}(\mathcal{Z},\mathcal{Z}') \in \mathbb{R}^{n \times m}\) is a matrix obtained by evaluating the scalar-valued kernel-based function \( \mathcal{O} \mathcal{O}' k(\bm{z},\bm{z}') \) on each pair \( (\bm{z},\bm{z}') \in \mathcal{Z} \times \mathcal{Z}' \), where \( \mathcal{Z} = \{ \bm{z}_i \}_{i=1}^n,\ \mathcal{Z}' = \{ \bm{z}'_j \}_{j=1}^m\), and \( \mathcal{O} \in \{\mathcal{L}, \mathcal{B}_c\} \), \( \mathcal{O}' \in \{\mathcal{L}', \mathcal{B}'_c\} \) are linear operators acting on \( z \) and \( z' \), respectively. Similarly, $\mathcal{O}\bm{k}(\bm{z},\mathcal{Z'}) \in \mathbb{R}^m $, $\bm{s}(\mathcal{Z'}) \in \mathbb{R}^m$ and $\bm{b}_c(\mathcal{Z'}) \in \mathbb{R}^m$ are vectors obtained by evaluating the functions $\mathcal{O}k(\bm{z},\bm{z}')$, $s(\bm{z}')$ and $b_c(\bm{z}')$ on every $\bm{z}' \in \mathcal{Z'}$.

\subsection{Kernel Gram matrix inversion with stochastic dual descent} \label{ss:kinv}
The Gram matrix $\mathbf{K}$, derived from $N$ collocation points, necessitates storage that scales quadratically, $\mathcal{O}(N^2)$. Moreover, a direct computation of the coefficient vector or the optimal representer weights $\tilde{\bm{\alpha}}_N = \mathbf{K}^{-1}g(\mathcal{Z}_{col})$, where $\tilde{\bm{\alpha}}_N \in \mathbb{R}^N$, for obtaining the posterior mean exhibits a cubic computational complexity with respect to the number of collocation points, i.e., $\mathcal{O}(N^3)$. Recognizing the essential role of the aforementioned coefficient computation in determining the posterior mean, and the requirement of an additional analogous computation in sampling from the GPPS posterior or obtaining the posterior covariance function, we propose the utilization of SDD \cite{lin2024stochastic} for obtaining representer weights.
The SDD methodology synergistically integrates three pivotal components to achieve efficient optimization. Firstly, it reformulates the optimization problem by minimizing its dual counterpart, denoted as
\begin{equation}
\bar{L}(\bm{\alpha}_N) = \frac{1}{2}\|\bm{\alpha}_N\|^2_{\mathbf{K} + \lambda \mathbf{I}} - \bm{\alpha}_N^T g(\mathcal{Z}_{col}),
\end{equation}
instead of the primal objective
\begin{equation}
L(\bm{\alpha}_N) = \frac{1}{2}\|g(\mathcal{Z}_{col}) - \mathbf{K}\bm{\alpha}_N\|_2^2 + \frac{\lambda}{2}\|\bm{\alpha}_N\|^2_{\mathbf{K}}.
\end{equation}
It is important to note that while both formulations share the same optimal solution, $\tilde{\bm{\alpha}}_N$, the dual objective exhibits a superiorly conditioned Hessian matrix,
\begin{equation}
\nabla^2 \bar{L}(\bm{\alpha}_N) = \mathbf{K} + \lambda \mathbf{I},
\end{equation}
in contrast to the primal Hessian,
\begin{equation}
\nabla^2 L(\bm{\alpha}_N) = \mathbf{K}(\mathbf{K} + \lambda \mathbf{I}).
\end{equation}
This advantageous property of the dual Hessian facilitates the employment of larger stable step-sizes, thereby accelerating the convergence rate. Secondly, the computational cost associated with gradient estimation is reduced from $\mathcal{O}(N^2)$ to $\mathcal{O}(N)$ through the adoption of randomized gradient estimation techniques, with random coordinate selection being a preferred strategy. Lastly, the algorithm incorporates Nesterov's momentum and geometric iterate averaging to further enhance the convergence characteristics of the optimization process. The procedural outline of the SDD algorithm is formally presented in Algorithm \ref{alg:sdd_GPPS}.

\begin{algorithm}[ht]
\caption{Stochastic dual descent for obtaining representer weights or $\tilde{\bm{\alpha}}_N$ in GPPS}
\label{alg:sdd_GPPS}
\begin{algorithmic}[1]
\Require Kernel Gram matrix $\mathbf{K} \in \mathbb{R}^{N \times N}$ with rows $\mathbf{k}_1, \dots, \mathbf{k}_N$, vector of source and boundary term evaluations $g(\mathcal{Z}_{col}) \in \mathbb{R}^{N}$, regularization parameter $\lambda > 0$, number of SDD iterations $I \in \mathbb{N}_{+}$, mini-batch size $\mathrm{b} \in \{1, \dots, k\}$, step size $\beta > 0$, momentum coefficient $\rho \in [0,1)$, averaging weight $r \in [0,1]$
\Ensure Approximation of the representer weights $\bar{\bm{\alpha}}_N \approx (\mathbf{K} + \lambda \mathbf{I})^{-1}g(\mathcal{Z}_{col})$

\State Initialize the velocity vector $\mathcal{V}^0 = \mathbf{0} \in \mathbb{R}^{N}$, the solution vector $\bm{\alpha}_{N}^0 = \mathbf{0} \in \mathbb{R}^{N}$, and the averaged solution vector $\bar{\bm{\alpha}}_{N}^0 = \mathbf{0} \in \mathbb{R}^{N}$
\For{$i = 1$ \textbf{to} $I$}  
    \State Sample a mini-batch $\mathcal{J}_i = \{j^{i}_1, \dots, j^{i}_{\mathrm{b}}\} \sim \text{Uniform}(\{1, \dots, k\})$
    \State Compute a stochastic estimate of the gradient:
    \[
    \mathcal{G}^i = \frac{N}{B} \sum_{j \in \mathcal{J}_i} \left(\left(\mathbf{K}_j + \lambda \mathbf{e}_j\right)^\top \left(\bm{\alpha}_{N}^{i-1} + \rho \mathcal{V}^{i-1}\right) - g(\mathcal{Z}_{col})_j\right) \mathbf{e}_j
    \]
    \State Update the velocity vector: $\mathcal{V}^i \gets \rho \mathcal{V}^{i-1} - \beta \mathcal{G}^i$
    \State Update the solution vector: $\bm{\alpha}_{N}^i \gets \bm{\alpha}_{N}^{i-1} + \mathcal{V}^i$
    \State Perform geometric averaging of the solution: $\bar{\bm{\alpha}}_{N}^i \gets r \bm{\alpha}_{N}^i + (1 - r) \bar{\bm{\alpha}}_{N}^{i-1}$
    \State (Optional) Compute the primal loss function:
    \[
    L^i(\bm{\alpha}_{N}) = \frac{1}{2} \|g(\mathcal{Z}_{col}) - \mathbf{K} \bm{\alpha}_{N}^i\|^2 + \frac{\lambda}{2} \|\bm{\alpha}_{N}^i\|_{\mathbf{K}}^2
    \]
\EndFor
\State \Return $\bar{\bm{\alpha}}_{N}^I$
\end{algorithmic}
\end{algorithm}
\subsection{$h$-adaptive probabilistic solver using clustering-based active learning}

In this section, we present an active learning strategy to efficiently determine optimal locations for collocation points $\mathcal{Z}_{col}$ when solving PDEs with GPPS. Considering the potentially substantial computational burden associated with solving PDEs numerically, a sequential active learning strategy that identifies a single optimal collocation point per iteration may prove inefficient. Consequently, we propose employing a sample-based active learning paradigm \cite{dubourg2013metamodel,zhang2019multi} wherein a set of collocation points is selected concurrently, thus facilitating the utilization of parallel computing resources.

The active learning procedure commences with the generation of an initial set of collocation points $\mathcal{Z}_{col}^{0}$, followed by the construction of a candidate pool of collocation points $\mathcal{Z}_{\text{pool}}$ sampled from an easy to sample and appropriate density. In the context of this article, we employ the Sobol sequence for the generation of the candidate pool and the initial collocation set. To mitigate the redundancy arising from the close proximity of candidate points to existing collocation points, and to enhance the accuracy of the GPPS over extended local regions, the candidate pool $\mathcal{Z}_{\text{pool}}$ is subjected to a filtering process based on an exclusion radius $r_{\text{excl}}$ to obtain the filtered set $\mathcal{Z}_{\text{filt}}$, which can be represented as 
\begin{equation}
    \mathcal{Z}_{\text{filt}} = \left\{ \bm{z} \in \mathcal{Z}_{\text{pool}} \;\middle|\; \min_{\bar{\bm{z}} \in \mathcal{Z}_i} \|\bm{z} - \bar{\bm{z}}\| \geq r_{\text{excl}} \right\}
\end{equation}
The points in $\mathcal{Z}_{\text{filt}}$ are then ranked according to a predefined acquisition or utility function. Subsequently, only a fraction ($f_{\text{ret}}$) of these filtered points, denoted as $\mathcal{Z}_{\text{ret}}$, are retained based on the descending order of their acquisition function values. In pursuit of variance reduction for this study, we adopt the posterior covariance, which serves as a quantitative measure of the uncertainty at a given location induced by the spatial discretization, as our acquisition function. The acquisition function can be expressed as
\begin{equation}
    \zeta(\tilde{\bm{z}}) =  k(\tilde{\bm{z}},\tilde{\bm{z}}) - \mathcal{A}(\tilde{\bm{z}})^T \mathbf{K}^{-1} \mathcal{A}(\tilde{\bm{z}})
    \label{eq:acq}
\end{equation}

The retained set of collocation points is then partitioned into $n_c$ clusters using the K-means clustering algorithm \cite{hartigan1979algorithm}. Following this, a representative collocation point is selected from each cluster according to the criterion:
\begin{equation}
    \bm{z}_j^* = \arg\min_{\bm{z} \in C_j} \|\bm{z} - \operatorname{centroid}(C_j)\|, \quad \text{for } j = 1, \dots, n_c
\end{equation}
where $\bm{z}_j^*$ represents the selected point from the $j$-th cluster $C_j$. This selection strategy inherently prevents the addition of closely spaced points within a single iteration. The set of selected points, denoted by $\mathcal{Z}_{\text{act}} = \{ \bm{z}_j^* \mid j = 1, \dots, n_c \}$, is then incorporated into the existing set of collocation points $\mathcal{Z}_{col}^{i-1}$ at the $i$-th iteration to yield the updated set $\mathcal{Z}_{col}^i$, formally expressed as $\mathcal{Z}_{col}^{i} \gets \mathcal{Z}_{col}^{i-1} \cup \mathcal{Z}_{\text{act}}$, which is then used to solve the PDE with GPPS. This iterative refinement process continues until the GPPS-based solution achieves a pre-specified level of approximation accuracy. It is pertinent to note that the retained fraction $f_{\text{ret}}$, the number of clusters $n_c$, and the cardinality of the candidate pool, $n_p$, constitute hyperparameters whose optimal values are contingent upon the specific PDE under consideration. However, as a general guideline, the product $f_{\text{ret}}n_p$ should significantly exceed the number of points acquired from a single active learning iteration, which is also equal to $n_c$.

\subsection{Posterior inference and sampling for GPPS}
Drawing upon the discussion in Section \ref{ss:kinv}, a supplementary yet analogous computational procedure is indispensable for the acquisition of samples from the GPPS predictive posterior distribution. This additional exigency arises from the use of the pathwise sampling algorithm \cite{wilson2021pathwiseconditioninggaussianprocesses}, which mandates the determination of optimal representer weights, denoted as $\tilde{\bm{\beta}}_N \in \mathbb{R}^N$, for obtaining the uncertainty reduction term. This term, representing the attenuation of predictive variance informed by the enforcement of the PDE $\mathcal{P}$ on the set of collocation points $\mathcal{Z}$, can be expressed as
\begin{equation}
\hat{\bm{m}}_{\bm{\beta}_N} (\tilde{\bm{z}}) = \mathcal{A}(\tilde{\bm{z}})^T \tilde{\bm{\beta}}_N.
\label{eq:urt}
\end{equation}
Here, the vector of optimal representer weights, $\tilde{\bm{\beta}}_N$ is obtained through the following relation
\begin{equation}
 \tilde{\bm{\beta}}_N = \left(\mathbf{K}^{-1} + \lambda \mathbf{I} \right) \mathcal{A}(\tilde{\bm{z}}),
 \label{eq:beta}
\end{equation}
 Note that the representer weights $\bm{\beta}_N$ are different from the representer weights $\bm{\alpha}_N$, which were used for the posterior mean computation. By integrating this uncertainty reduction term with the predictive mean and samples drawn from the prior distribution, the pathwise sampling methodology facilitates the generation of predictive posterior samples. This process is succinctly captured by the subsequent expression
\begin{equation}
\left(\tilde{u}|\mathcal{P}\right)(\tilde{\bm{z}}) = \underbrace{u(\tilde{\bm{z}})}_{\text{prior sample}} + \underbrace{\hat{\bm{m}}_{\tilde{u}|\mathcal{P}}(\tilde{\bm{z}})}_{\text{predictive mean}} -\underbrace{\hat{\bm{m}}_{\bm{\beta}_N}(\tilde{\bm{z}})}_{\text{uncertainty reduction term}},
\label{eq:psamp}
\end{equation}
where $u \sim  \mathcal{GP}(0, k(\cdot,\cdot))$. The algorithmic specifics of using pathwise sampling for generating samples from the predictive posterior distribution of GPPS are given in Algorithm \ref{alg:GPPS_pathwise_inference}.
\begin{algorithm}
\caption{Inference via pathwise sampling for GPPS}
\label{alg:GPPS_pathwise_inference}
\begin{algorithmic}[1]
\Require Test input $\tilde{\bm{z}}$, kernel Gram matrix $\mathbf{K}$, boundary and source term vector $g(\mathcal{Z}_{col}) \in \mathbb{R}^{N}$, prior sample $u(\cdot) \sim \mathcal{GP}(0, k(\cdot, \cdot))$, and learned representer weights $\tilde{\bm{\alpha}}_N$
\Ensure Realizations from the predictive posterior distribution of the GPPS

\State Compute the predictive posterior mean using learned representers:
$
\hat{\bm{m}}_{u|\mathcal{P}}(\tilde{\bm{z}}) = \mathcal{A}(\tilde{\bm{z}})^\top \tilde{\bm{\alpha}}_N
$

\State Evaluate the uncertainty correction term via new representer weights $\tilde{\bm{\beta}}_N$:
$
\hat{\bm{m}}_{\bm{\beta}_N}(\tilde{\bm{z}}) = \mathcal{A}(\tilde{\bm{z}})^\top \tilde{\bm{\beta}}_N
$

\State Compute the sample from the predictive posterior distribution:
$
\left(\tilde{u}|\mathcal{P}\right)(\tilde{\bm{z}}) = u(\tilde{\bm{z}}) + \hat{\bm{m}}_{u|\mathcal{P}}(\tilde{\bm{z}}) - \hat{\bm{m}}_{\bm{\beta}_N}(\tilde{\bm{z}})
$

\State \Return Predictive posterior sample $\left(\tilde{u}|\mathcal{P}\right)(\tilde{\bm{z}})$
\end{algorithmic}
\end{algorithm}

\section{Numerical Examples}\label{Numercial_eg}
In this section, we evaluate the performance of the proposed scalable $h$-adaptive probabilistic solver, leveraging SDD-based optimization and sample-based active learning, across three representative case studies. The selected numerical examples comprise standard benchmark problems, including Poisson's equation in two and three-dimensional domains and a one-dimensional time-dependent heat equation solved in a space-time formalism. 
These numerical examples underscore both the efficiency of SDD optimization and the effectiveness of the active learning strategy, which iteratively selects the most informative collocation points and reduces error based on some acquisition function (e.g., variance reduction). Depending on the complexity of the problem and the number of clusters employed (typically ranging from 5 to 20), convergence is achieved within 10 to 20 iterations for the examples considered. Table~\ref{table:compariosn_cs} provides a comparative analysis of various numerical case studies across different frameworks by reporting the relative mean squared error (MSE) for a fixed number of collocation points. The reported errors demonstrate the superior performance of the proposed solver in comparison to other approaches. 
Furthermore, throughout the study, we utilize a stationary covariance kernel for GPPS, given by
\begin{equation}
    k(\bm{x}, \bm{x}') = s^2 \exp \left( -\frac{1}{2} \sum_{r=1}^{d} \frac{(x_r - x_r')^2}{\ell_r^2} \right),
\end{equation}
where $s$ represents the signal strength, and $\ell_r$ $(\text{for }r=1,\dots,d)$ denote the characteristic length scales in each spatial dimension. 
\begin{table}[ht!]
\centering
    \begin{threeparttable}
        \caption{Comparison of relative MSE for different numerical examples among GP with SDD and clustering-based active learning (SDD GP (AL)), GP with direct inversion of kernel Gram matrix (Exact GP), and Physics-informed neural networks (PINNs). }
        \label{table:compariosn_cs}
        \begin{tabular}{lccc}
        \hline
        \multirow{2}{*}{\textbf{Case studies}} & \multicolumn{3}{c}{\textbf{Frameworks}} \\ \cline{2-4} 
        & SDD GP (AL) & Exact GP & PINNs \\
        \midrule
        Poisson's equation on a circular disk & $\mathbf{1.21} \%$ & 1.83 \% & 83.91\% \\
        Poisson's equation in three dimensions & $\mathbf{1.41}\%$ &  2.12\% & 206.52\%\\
        Time-dependent heat equation in one dimension & $\mathbf{1.11}\%$ & 3.71\% & 12.80\%\\
        \bottomrule
        \end{tabular}
    \end{threeparttable}
\end{table}

\subsection{Case Study 1: Poisson equation on a circular disk}
In this first case study, we focus on solving the Poisson equation, which is a second-order elliptic partial differential equation. This equation arises in many fields, such as electrostatics and fluid dynamics. The problem is posed on a unit disk, defined by the domain $\Omega = \{ \bm{x}: ||\bm{x}||^2 \leq 1 \}$, with the governing equation given as:
\begin{equation}\label{eq:poisson2d}
\begin{aligned}
    -\nabla^2 u(\bm{x}) = 1, \quad \forall \bm{x} \in \Omega, \\
     u(\bm{x}) = 0, \quad \forall \bm{x} \in \partial \Omega
\end{aligned}
\end{equation}
The analytical solution for Eq. \eqref{eq:poisson2d} is given by:
\begin{equation}
\begin{aligned}
    u(\bm{x}) = \frac{1 - ||\bm{x}||^2}{4}.
    \label{eq:sol_pois2d}
\end{aligned}
\end{equation}
\begin{figure}[ht!]
    \centering
    \includegraphics[width=\linewidth]{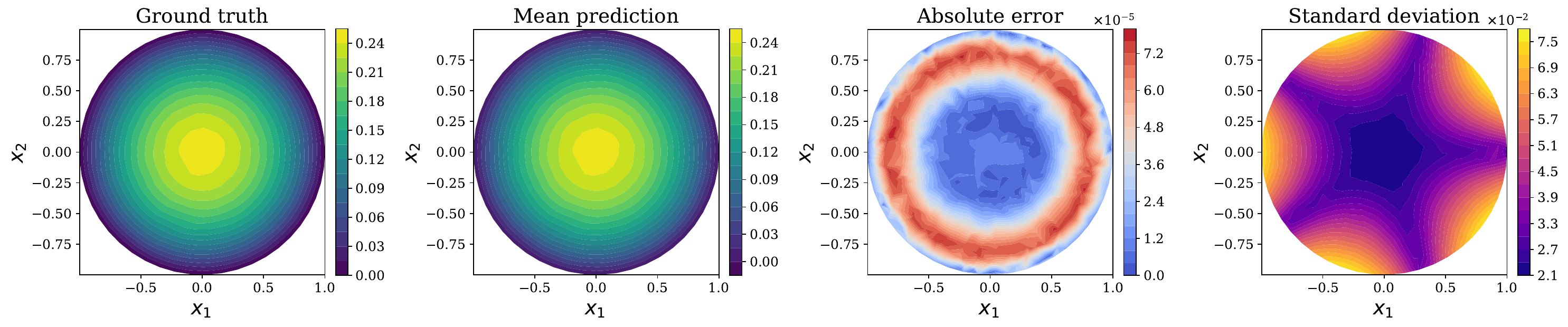}
    \caption{The analytical ground truth, the posterior mean and the standard deviation, as well as the absolute error between the posterior mean and the analytical ground truth, obtained using the GPPS with direct inversion of the kernel Gram matrix for the Poisson equation on a circular disk.}
    \label{fig:gp_poisson_2d}
\end{figure}
\begin{figure}[ht!]
    \centering
    \includegraphics[width=\linewidth]{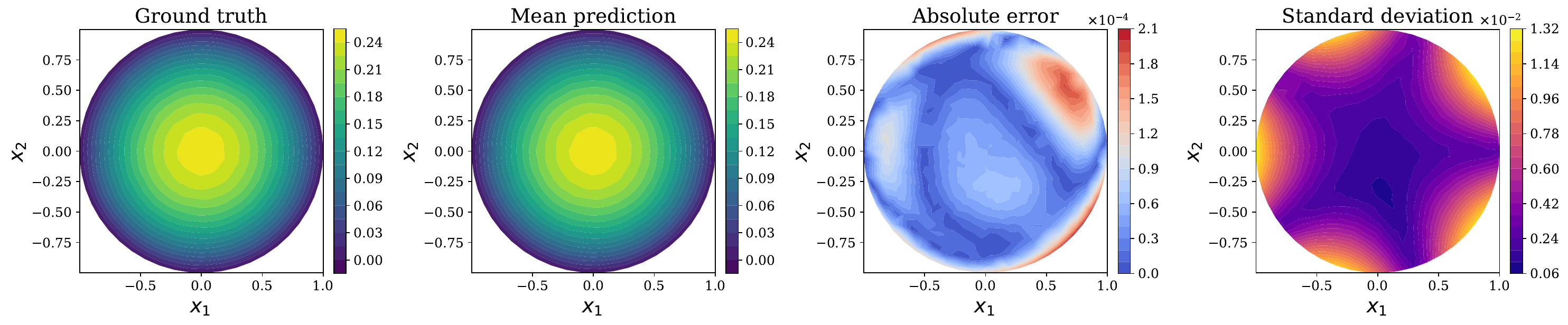}
    \caption{The analytical ground truth, the posterior mean, the standard deviation, and the absolute error between the posterior mean and the analytical ground truth, obtained using GPPS with SDD, without the application of active learning, for the Poisson equation on a circular disk.}
    \label{fig:gp_poisson_2d_sdd}
\end{figure}
\begin{figure}[ht!]
    \centering
    \includegraphics[width=0.95\linewidth]{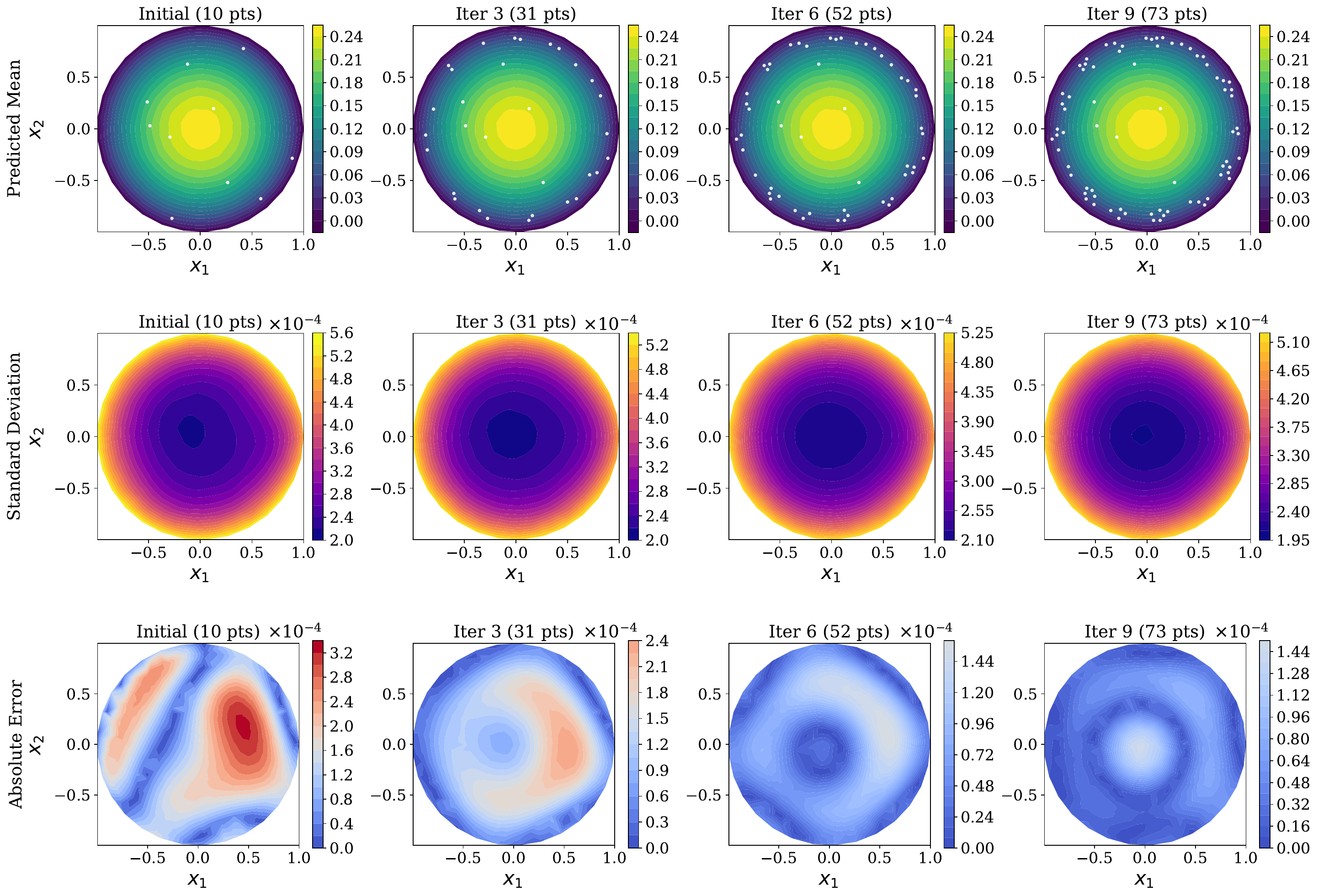}
    \caption{Variance reduction achieved through clustering-based active learning for the Poisson equation on a circular disk. The figure presents the evolution of the posterior mean, posterior standard deviation, and the absolute error between the posterior mean and analytical ground truth. From left to right, as the number of sampled collocation points increases, both variance and error decrease.}
    \label{fig:poisson_2d_al_cl}
\end{figure}
\begin{figure}[ht!]
    \centering
    \includegraphics[width=1.0\linewidth]{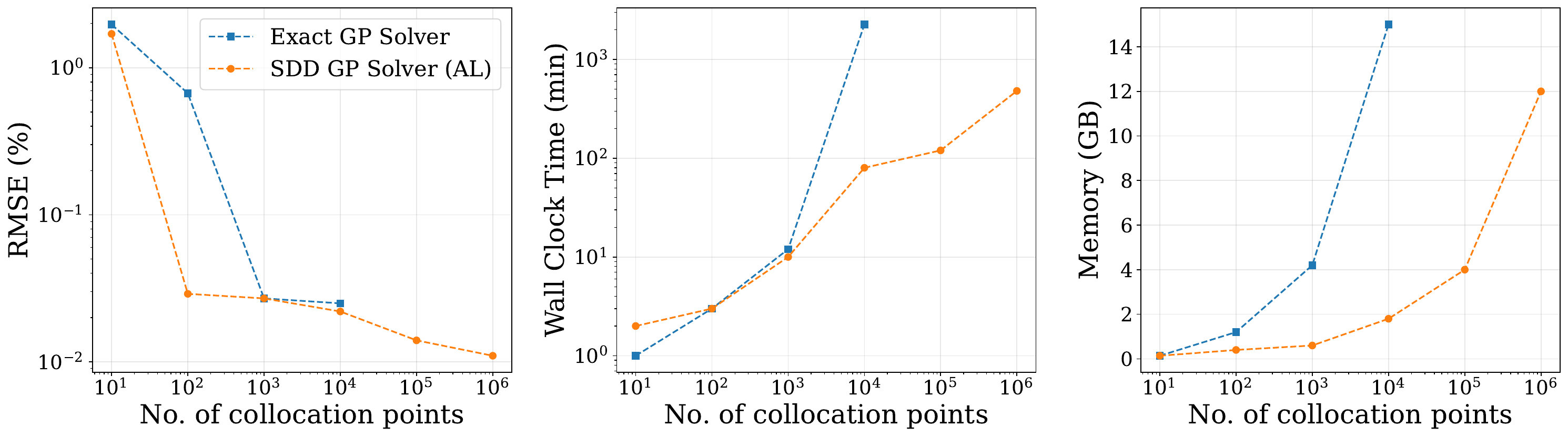}
    \caption{ Scalability and performance comparison between GPPS with direct kernel Gram matrix inversion and GPPS with SDD and clustering-based active learning for the Poisson equation on a circular disk. Note that the active learning algorithm is initialized with five collocation points, and the data point representing ten collocation points in the plots corresponds to the state after the first active learning iteration. The \textbf{left}, \textbf{middle}, and the \textbf{right} panels depict the variation of relative-MSE with collocation points, the wall clock time, and the total memory usage with collocation points, respectively.}
    
    \label{fig:poisson2D_scalabilty}
\end{figure}
%
The solution and its deviation from the reference solution in Eq. \eqref{eq:sol_pois2d}, as well as the posterior standard deviation, obtained using GPPS with direct inversion of the kernel Gram matrix (exact GPPS), are presented in Fig.~\ref{fig:gp_poisson_2d}. Fig.~\ref{fig:gp_poisson_2d_sdd} presents the results when SDD was used instead of direct inversion. Note that for the solver variants presented so far, active learning was not used. Fig.~\ref{fig:poisson_2d_al_cl} showcases the solution accuracy and uncertainty estimates obtained using the proposed $h$-adaptive scalable probabilistic solver, i.e., GPPS with SDD and clustering-based active learning. The figure also illustrates the iterative improvement of the solution and reduction of posterior standard deviation as the number of selected collocation points increases over active learning iterations. For this example, the active learning procedure effectively identified $n_i =60$ internal and $n_b =13$ boundary discretization points, at which the desired error tolerance was satisfied. Additionally, although the same number of boundary and internal points was used to generate Fig.~\ref{fig:gp_poisson_2d} and Fig.~\ref{fig:gp_poisson_2d_sdd}, they were randomly generated. Furthermore, we put the claimed scalability of the proposed approach to the test, with findings presented in Figure~\ref{fig:poisson2D_scalabilty}. This analysis reveals the better scaling properties of the proposed probabilistic framework in comparison to the exact GPPS, primarily owing to its ability to achieve lower errors with the same number of collocation points, a reduced memory footprint, and accelerated solution times. It was observed that the exact GPPS for solving the PDE demanded around nineteen hours of wall-clock time when utilizing $10^4$ collocation points. This contrasts sharply with the GPPS employing SDD and a clustering-based active learning approach, which reduced the computational time to around one and a half hours. Due to the computational infeasibility on the hardware used, the wall-clock times for the exact GPPS with $10^5$ and $10^6$ collocation points were not reported. It is also worth noting that solving the PDE under consideration using $10^6$ collocation points with exact GPPS will be impractical even on high-end computing hardware.

\subsection{Case Study 2: Poisson equation in three dimensions}
For this second case study,  we consider the Poisson equation in a three-dimensional domain $\Omega = [0,1]^3 \subset \mathbb{R}^3$, which is mathematically represented as

\begin{equation} \label{eq:poisson3d}
\begin{aligned}
-\nabla^2 u(\bm{x}) &= f(\bm{x}), \quad \bm{x} \in \Omega, \\
f(\bm{x}) &= 3\pi^2 \sin(\pi x_1) \sin(\pi x_2) \sin(\pi x_3), \\
u(\bm{x}) &= 0, \quad \text{for} \quad \bm{x} \in \partial\Omega.
\end{aligned}
\end{equation}

\noindent Here, $u(\bm{x})$ denotes the unknown scalar field, while $f(\bm{x})$ serves as a smooth, spatially varying forcing function. The problem is completed by the imposition of homogeneous Dirichlet boundary conditions, specifying $u(\bm{x})=0$ for all $\bm{x}$ on the boundary $\partial\Omega$ of the computational domain. This particular choice of forcing function and boundary conditions also admits an analytical solution.\par
\begin{figure}[ht!]
    \centering
    \includegraphics[width=\linewidth]{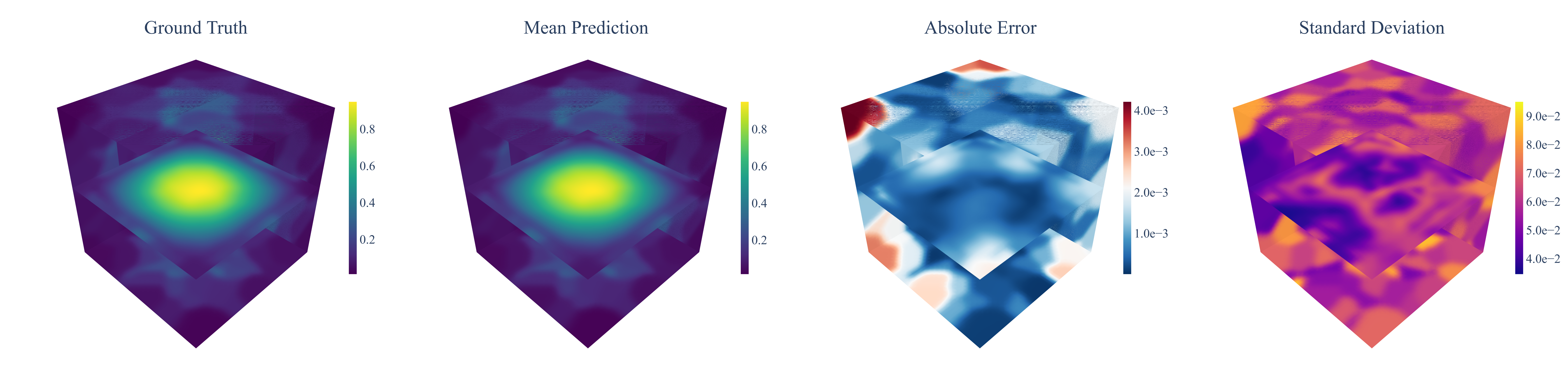}
    \caption{The analytical ground truth, the posterior mean, the standard deviation, and the absolute error between the posterior mean and the analytical ground truth, obtained using GPPS with SDD, without the application of active learning, for the three-dimensional Poisson equation.}
    \label{fig:3d_poisson_gp_sdd}
\end{figure}
\begin{figure}[ht!]
    \centering
    \includegraphics[width=\linewidth]{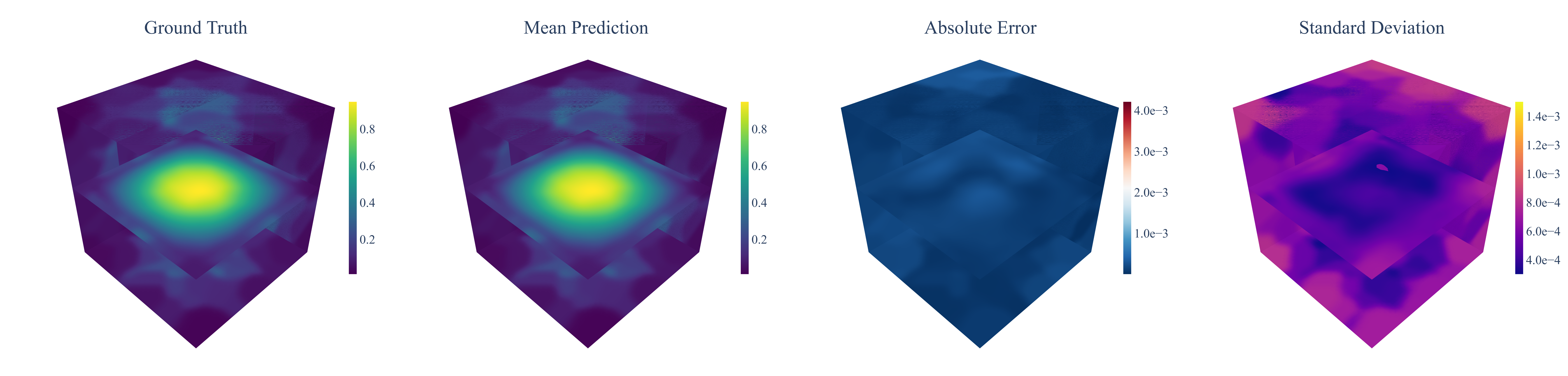}
    \caption{ The analytical ground truth, the posterior mean, the standard deviation, and the absolute error between the posterior mean and the analytical ground truth, obtained using GPPS with SDD and clustering-based active learning, for the three-dimensional Poisson equation.}
    \label{fig:3d_poisson_cl}
\end{figure}
\begin{figure}[ht!]
    \centering
    \begin{subfigure}[t]{0.45\linewidth}
        \centering
        \includegraphics[width=\linewidth]{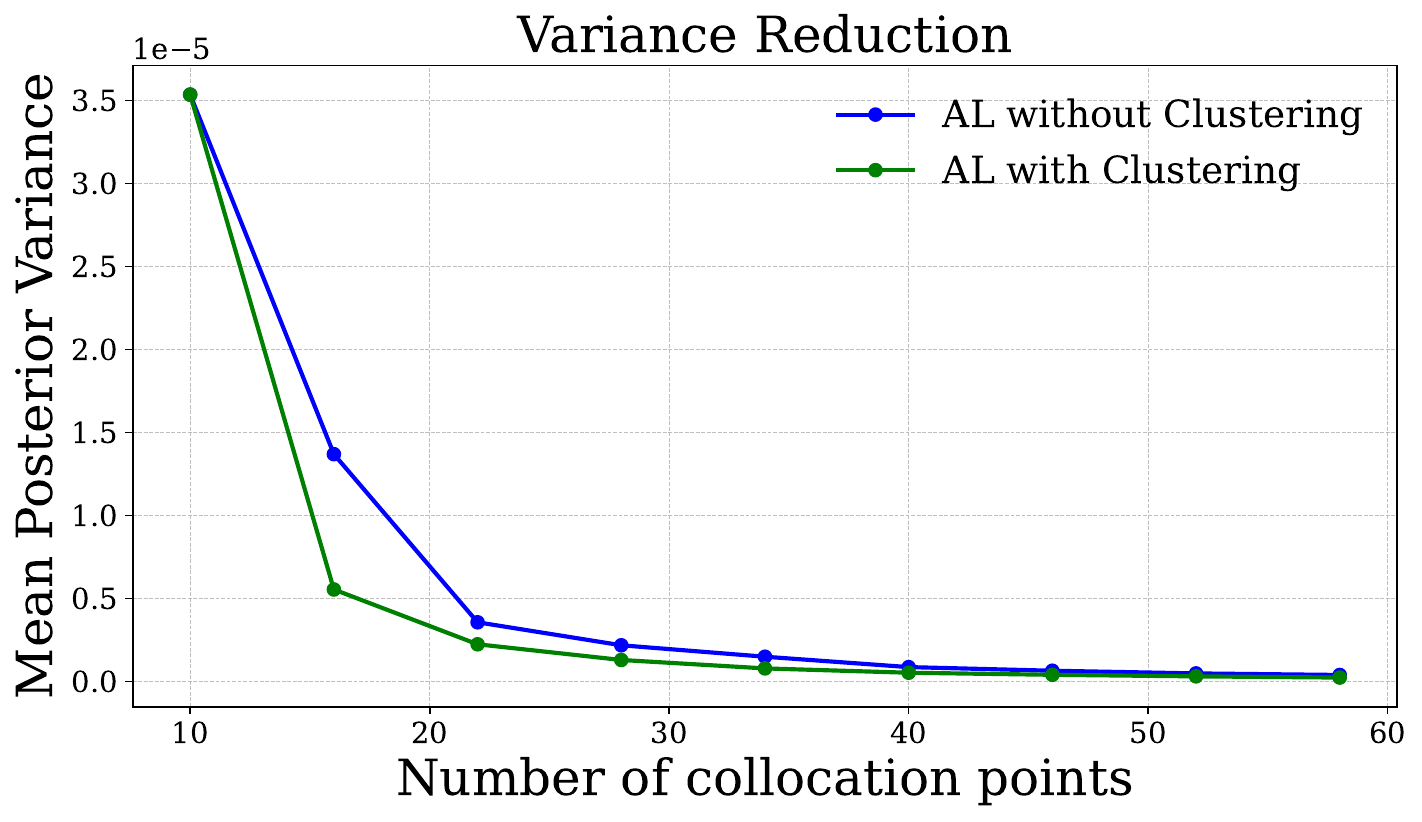}
        \caption{}
    \end{subfigure}
    \hfill
    \begin{subfigure}[t]{0.45\linewidth}
        \centering
        \includegraphics[width=\linewidth]{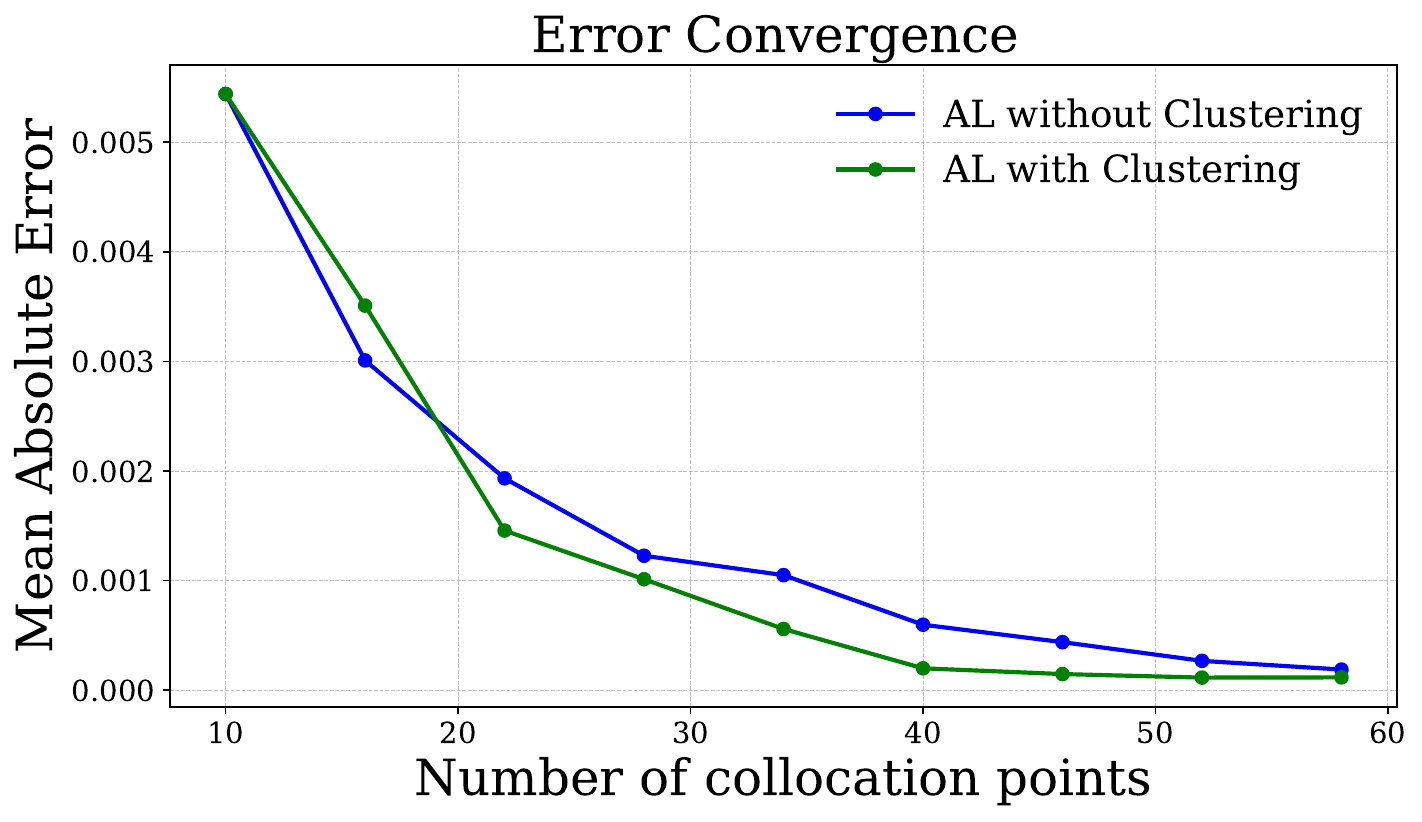}
        \caption{}
        \label{fig:error_conv}
    \end{subfigure}
    \caption{(a) Variance and (b) error convergence for the three-dimensional Poisson equation using GPPS with SDD and active learning. Two variants of active learning are presented: one leveraging clustering and the other operating without it.}
    \label{fig:conv_results}
\end{figure}
\begin{figure}[ht!]
    \centering
    \includegraphics[width=\linewidth]{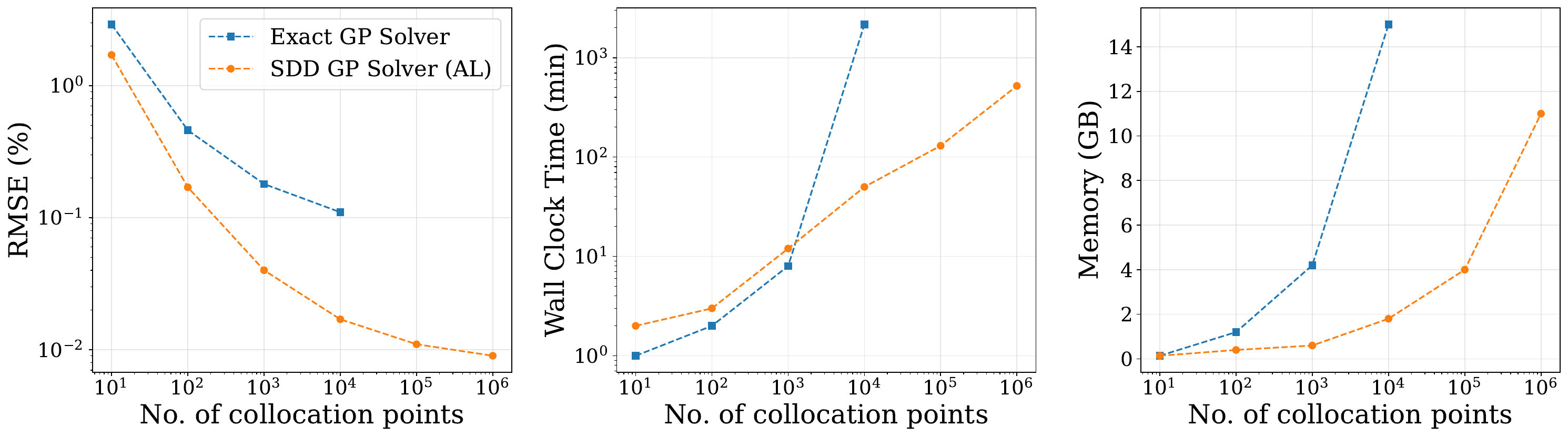}
    \caption{Scalability and performance comparison between GPPS with direct kernel Gram matrix inversion and GPPS with SDD and clustering-based active learning for the three-dimensional Poisson equation. The \textbf{left}, \textbf{middle}, and the \textbf{right} panels depict the variation of relative-MSE with collocation points, the wall clock time, and the total memory usage with collocation points, respectively.}
    \label{fig:poisson3D_scalabilty}
\end{figure}

Firstly, we present the results, predicted posterior mean and standard deviation, obtained from the solver which couples GP and SDD but does not employ active learning, and compare them against the analytical ground truth in Figure~\ref{fig:3d_poisson_gp_sdd}. Subsequently, two distinct variants of the SDD solver, augmented with active learning capabilities, are assessed. Specifically, one variant employs a clustering-based approach for optimal collocation point selection, while the other proceeds without such a strategy. Figure \ref{fig:conv_results} presents a comparative analysis of these two approaches, elucidating their respective convergence behaviors. The figure demonstrates a reduction in both total variance and mean absolute error with an increasing number of collocation points, while simultaneously showcasing the advantageousness of the clustering approach over the non-clustering alternative. Furthermore, the outcomes from GPPS with SDD and clustering-based active learning are presented in Figure~\ref {fig:3d_poisson_cl}. Lastly, scalability analyses, analogous to those performed in the preceding case study, are also undertaken for this problem, with the corresponding results presented in Figure~\ref{fig:poisson3D_scalabilty}. Similar trends to those observed in the previous case study are also seen in these analyses.

\subsection{Case Study 3: One-dimensional time-dependent heat equation}

As our final case study, we consider a one-dimensional time-dependent heat equation with a constant thermal diffusivity. The governing equation, along with the corresponding initial and boundary conditions, can be expressed as \\
\begin{equation} \label{eq:1dHeat}
\begin{aligned}
\frac{\partial u}{\partial t}(x,t) &= \alpha \frac{\partial^2 u}{\partial x^2}(x,t),\\
u(x, 0) = \sin(\pi x), \quad u(0, t) &= 0, \quad u(1, t) = 0, \quad \frac{\partial u}{\partial t}(x, 1) = 0. \\
\end{aligned}
\end{equation}
\noindent Here, \( u(x, t) \) represents the temperature at any spatiotemporal location $(x,t) \in \Omega_x \times \Omega_t = [0,1] \times [0,1] $, and \( \alpha = 0.01 \) is the thermal diffusivity coefficient.\par

\begin{figure}[ht!]
    \centering
    \includegraphics[width=\linewidth]{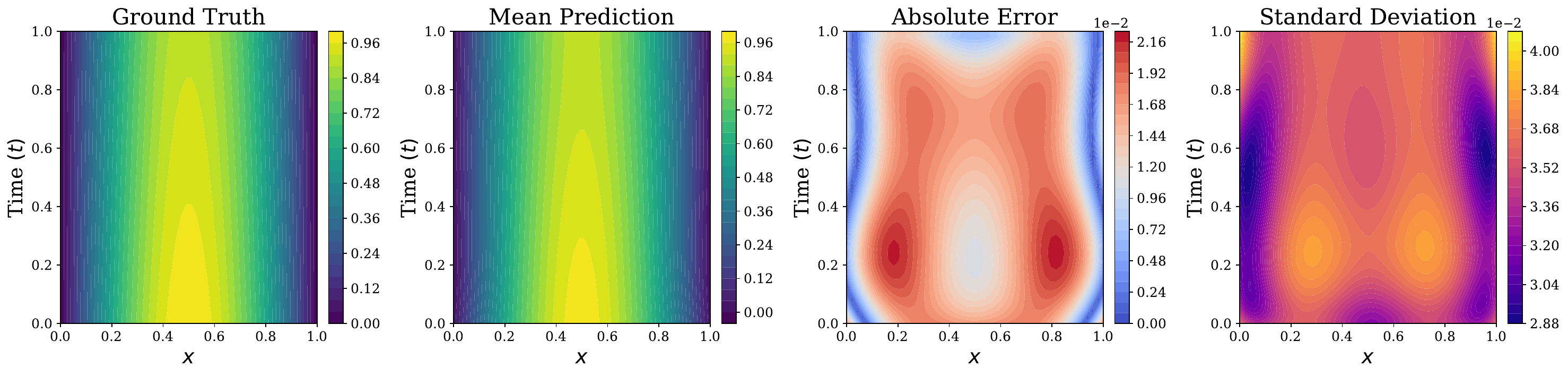}
    \caption{The ground truth, the posterior mean, the standard deviation, and the absolute error between the posterior mean and the ground truth, obtained using GPPS with SDD, without the application of active learning, for the one-dimensional time-dependent heat equation.}
    \label{fig:sdd_heat_al}
\end{figure}

\begin{figure}[ht!]
    \centering
    \includegraphics[width=\linewidth]{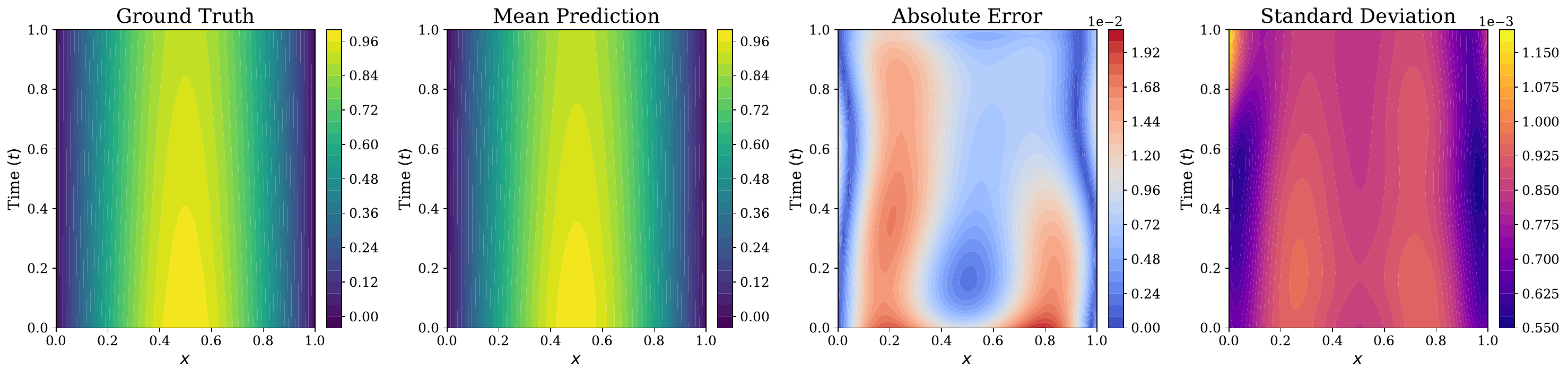}
    \caption{The ground truth, the posterior mean, the standard deviation, and the absolute error between the posterior mean and the ground truth, obtained using GPPS with SDD and clustering-based active learning, for the one-dimensional time-dependent heat equation.}
    \label{fig:heat_al_cluster}
\end{figure}

\begin{figure}[ht!]
    \centering
    \includegraphics[width=\linewidth]{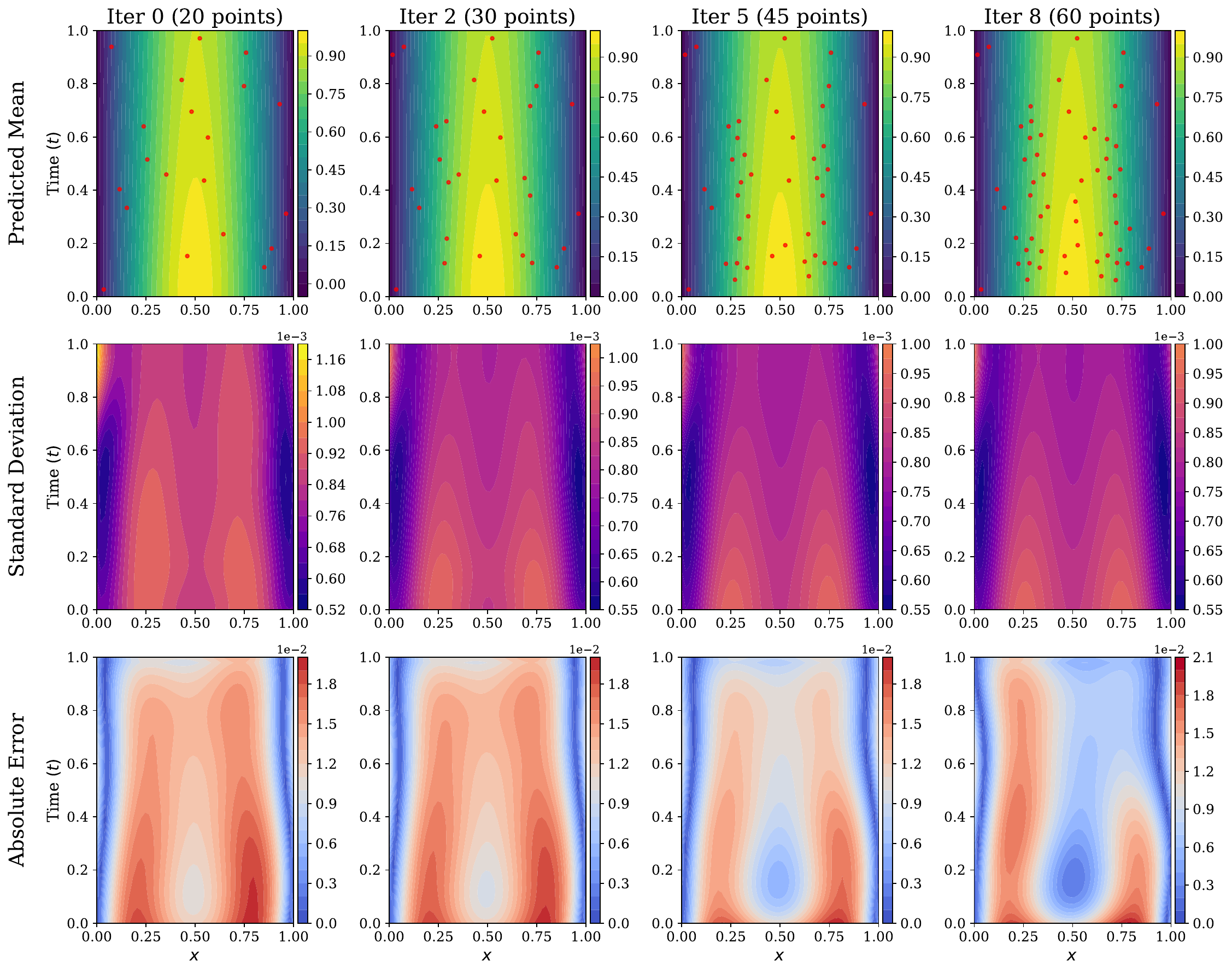}
    \caption{Variance reduction achieved through clustering-based active learning for the one-dimensional time-dependent heat equation. The figure presents the evolution of the posterior mean, posterior standard deviation, and the absolute error between the posterior mean and ground truth. From left to right, as the number of sampled collocation points increases, both variance/standard deviation and error decrease.}
    \label{fig:heat_al_iter}
\end{figure}

\begin{figure}[ht!]
    \centering
    \includegraphics[width=\linewidth]{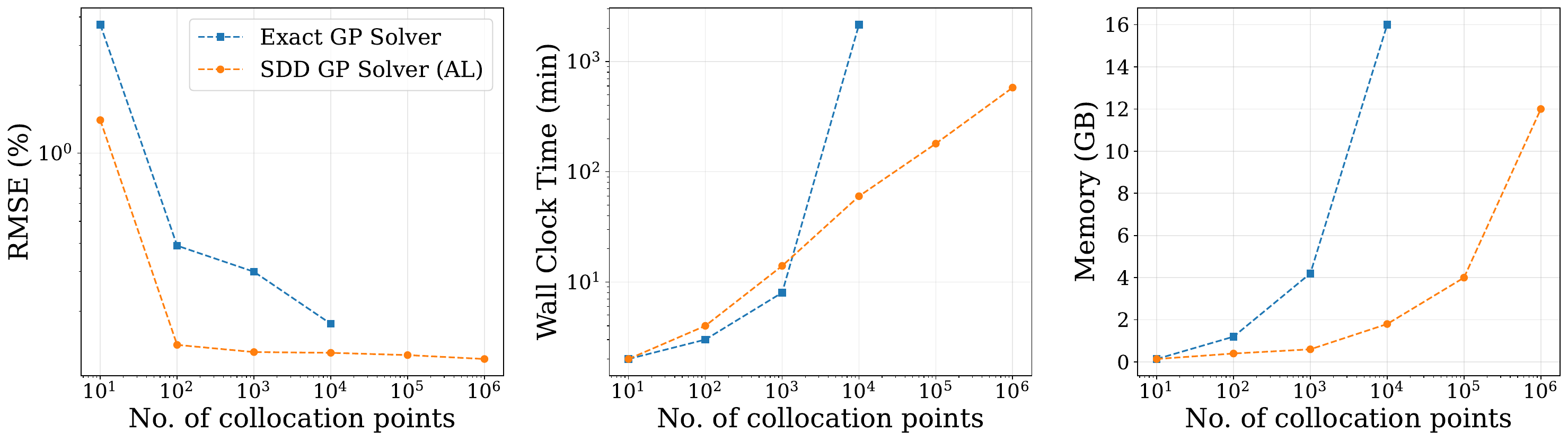}
    \caption{Scalability and performance comparison between GPPS with direct kernel Gram matrix inversion and GPPS with SDD and clustering-based active learning for the one-dimensional time-dependent heat equation. The \textbf{left}, \textbf{middle}, and the \textbf{right} panels depict the variation of relative-MSE with collocation points, the wall clock time, and the total memory usage with collocation points, respectively.}
    \label{fig:heat1D_scalabilty}
\end{figure}

Results obtained from the GPPS coupled with SDD, without the inclusion of active learning, are presented in Figure~\ref{fig:sdd_heat_al}; these specifically detail the predicted posterior mean, the absolute error with respect to the numerical solution, which is considered the ground truth, and the corresponding standard deviation. Furthermore, the performance of the proposed framework, which integrates the GPPS with SDD and clustering-based active learning, is illustrated through the Figures~\ref{fig:heat_al_cluster} and~\ref{fig:heat_al_iter}. Figure~\ref{fig:heat_al_iter}, in particular, showcases the evolution of posterior mean and standard deviation across successive active learning iterations, demonstrating a reduction in both prediction error and uncertainty with an increasing number of selected collocation points. Similar to prior investigations, scalability for this case study is also assessed, with the results provided in Figure~\ref{fig:heat1D_scalabilty}. 


\section{Conclusion}\label{Conclusion}
In this work, we introduced an $h$-adaptive and scalable GP-based probabilistic solver designed to address the inherent limitations of traditional GP-based PDE solvers, primarily their prohibitive cubic computational cost with increasing collocation points. The proposed approach leverages the SDD algorithm to reformulate the GP regression problem in the dual space, achieving a reduction in per-iteration computational cost from cubic to linear with respect to the number of collocation points, thereby enabling the solution of large-scale problems through GPs. Complementing this, we employ a clustering-based active learning strategy with a variance-based acquisition function for the careful placement of collocation points in regions of uncertainty, thus improving accuracy while being efficient. The efficacy and robustness of our proposed framework are assessed through numerical experimentation across a suite of benchmark PDEs, encompassing both elliptic and parabolic cases, formulated in spatial and spatio-temporal domains, respectively. Across the different PDE case studies, our findings consistently demonstrate that the presented methodology not only yields accurate posterior mean estimates but, crucially, also furnishes uncertainty quantification. Moreover, we perform scalability studies across all PDE case studies. These studies demonstrate that our methodology achieves scalability by mitigating memory consumption and circumventing the prohibitive time complexity typically associated with exact inference in conventional GP solvers.

\section*{Acknowledgements}
SK acknowledges the support received from the Ministry of Education (MoE) in the form of a Research Fellowship. SC acknowledges the financial support received from Anusandhan National Research Foundation (ANRF) via grant no. CRG/2023/007667. 
\section*{Code availability}
On acceptance, all the source codes to reproduce the results in this study will be made available to the public on GitHub by the corresponding author.


\end{document}